\documentclass{ecai}
\pdfoutput=1
\usepackage{times}
\usepackage{helvet}
\usepackage{courier}
\usepackage{subfig}
\usepackage{graphicx}

\usepackage{amssymb,amsmath,latexsym}
\usepackage{amsthm}

\usepackage{graphicx}

\usepackage{xcolor}

\usepackage{algorithm}
\usepackage{algorithmicx}
\usepackage[noend]{algpseudocode}
\algrenewcommand{\algorithmiccomment}[1]{\textcolor{gray}{{//\tt #1}}}

\newtheorem{theorem}{Theorem}

\newtheorem{claim}[theorem]{Claim}
\newtheorem{definition}[theorem]{Definition}

\usepackage{enumitem}

\newcommand{\gdp}{\ensuremath{\mathsf{GDP}}}
\newcommand{\hopgdp}{\ensuremath{\mathsf{HOpGDP}}}
\newcommand{\godel}{\ensuremath{\mathsf{GoDeL}}}

\newcommand{\gdpopt}{\hopgdp}
\newcommand{\gdpblind}{\ensuremath{\hopgdp_\mathsf{blind}}}
\newcommand{\hgnlmheur}{\ensuremath{h_{HL}}}
\newcommand{\classicallmheur}{\ensuremath{h_{L}}}
\newcommand{\astarlm}{\ensuremath{\text{A*-}\classicallmheur}}
\newcommand{\hgnlmgen}{\ensuremath{\mathsf{computeHGNLandmarks}}}
\newcommand{\addlm}{\ensuremath{\mathsf{addLM}}}
\newcommand{\addlmandorder}{\ensuremath{\mathsf{addLMandOrdering}}}
\newcommand{\unreachedlm}{\ensuremath{\mathsf{Unreached}}}
\newcommand{\acceptedlm}{\ensuremath{\mathsf{Accepted}}}
\newcommand{\requiredagainlm}{\ensuremath{\mathsf{ReqAgain}}}

\newcommand{\lmgenclassical}{\ensuremath{\mathsf{LMGEN}_C}}
\newcommand{\cpdomain}{\ensuremath{D_{classical}}}

\newcommand{\frombody}[3]{\noindent{\color{#1}{{$\bf [\![\!\![$}\underline{\scshape{#2}} {\scshape says:} {\slshape #3}{$\bf ]\!\!]\!]$}}}}
\renewcommand{\frombody}[3]{}

\newcommand{\mypar}[1]{\medskip\textbf{#1}.}
\newcommand{\pre}{\textrm{precond}}
\newcommand{\eff}{\textrm{effects}}
\newcommand{\head}{\textrm{head}}
\newcommand{\sub}{\textrm{subgoals}}
\newcommand{\network}{\textrm{network}}

\newcommand{\goal}{\textrm{goal}}

\newcommand{\cost}{\ensuremath{\mathsf{cost}}}
\newcommand{\sol}{\ensuremath{\mathcal{S}}}

\newcommand{\openlist}{\ensuremath{\mathsf{open}}}
\newcommand{\searchSpace}{\ensuremath{\mathsf{searchSpace}}}
\newcommand{\getSuccessors}{\ensuremath{\mathsf{getSuccessors}}}
\newcommand{\failure}{\ensuremath{\mathsf{failure}}}

\title{Cost-Optimal Algorithms\\ for Planning with Procedural Control Knowledge}
\author{
Vikas Shivashankar\institute{Knexus Research Corporation, National Harbor, MD, vikas.shivashankar@knexusresearch.com}\and
Ron Alford\institute{MITRE, McLean, VA, ralford@mitre.org} \and
Mark Roberts\institute{NRC Postdoctoral Fellow, Naval Research Laboratory, Washington DC, mark.roberts.ctr@nrl.navy.mil} \and
David W. Aha\institute{Naval Research Laboratory, Washington DC, david.aha@nrl.navy.mil}
}

\ecaisubmission

\begin{document}
\maketitle

\begin{abstract}
There is an impressive body of work on developing heuristics and other reasoning algorithms to guide search in optimal and anytime planning algorithms for classical planning. However, very little effort has been directed towards developing analogous techniques to guide search towards high-quality solutions in hierarchical planning formalisms like HTN planning, which allows using additional domain-specific procedural control knowledge. In lieu of such techniques, this control knowledge often needs to provide the necessary search guidance to the planning algorithm, which imposes a substantial burden on the domain author and can yield brittle or error-prone domain models. We address this gap by extending recent work on a new hierarchical goal-based planning formalism called Hierarchical Goal Network (HGN) Planning to develop the \textit{\textbf{H}ierarchically-\textbf{Op}timal \textbf{G}oal \textbf{D}ecomposition \textbf{P}lanner (\hopgdp)}, an HGN planning algorithm that computes \textit{hierarchically-optimal} plans. \hopgdp\ is guided by \hgnlmheur, a new HGN planning heuristic that extends existing admissible landmark-based heuristics from classical planning to compute admissible cost estimates for HGN planning problems.
Our experimental evaluation across three benchmark planning domains shows that \hopgdp\ 
compares favorably to
both optimal classical planners due to its ability to use domain-specific procedural knowledge, and a blind-search
version of \hopgdp\ due to the search guidance provided by \hgnlmheur.

\end{abstract}

\section{Motivation and Background}


Formalisms for automated planning (to represent and solve planning problems) broadly fall into either \textit{domain-independent planning} or \textit{domain-configurable planning}. Domain-independent planning formalisms, such as \textit{classical planning} requires that the users only provide models of the base actions executable in the domain. In contrast, domain-configurable planning formalisms (e.g., \textit{Hierarchical Task Network (HTN) planning}) allow users to supplement action models with additional domain-specific knowledge structures that increases the expressivity and scalability of planning systems.


An impressive body of work exploring search heuristics has been developed for classical planning that has helped speed up generation of high-quality solutions.
More specifically, search heuristics such as the relaxed planning graph heuristic~\cite{hoffmann2001ff}, landmark generation algorithms~\cite{hoffmann04landmarks,richter10lama}, and landmark-based heuristics~\cite{richter10lama,karpas09costoptimal} dramatically improved optimal and anytime planning algorithms by guiding search towards (near-) optimal solutions to planning problems.


Yet relatively little effort has been devoted to develop analogous techniques to guide search towards high-quality solutions in domain-configurable planning systems. In lieu of such search heuristics, domain-configurable planners often require additional domain-specific knowledge to provide the necessary search guidance. This requirement not only imposes a significant burden on the user, but also sometimes leads to brittle or error-prone domain models. In fact, getting the best of heuristic search and hierarchical procedural knowledge (to decomposes planning tasks) has remained an unsolved problem since planning competitions first focused on heuristic search at AIPS-98~\cite{long00ipc98}.


In this paper, we address this gap by developing the \textit{\textbf{H}ierarchically-\textbf{Op}timal \textbf{G}oal \textbf{D}ecomposition \textbf{P}lanner (\hopgdp)}, a hierarchical planning algorithm that uses admissible heuristic estimates to generate \textit{hierarchically-optimal} plans (i.e., plans that are valid and optimal with respect to the given hierarchical knowledge). \hopgdp\ leverages recent work on a new hierarchical planning formalism called \textit{Hierarchical Goal Network} (HGN) Planning \cite{shivashankar2012hierarchical,shivashankar2013godel}, which combines the hierarchical structure of HTN planning with the goal-based nature of classical planning.


In particular, our contributions are as follows:
\begin{itemize}
    \item \textbf{Admissible Heuristic}: We present an HGN planning heuristic --  \hgnlmheur (\textbf{H}GN \textbf{L}andmark heuristic) -- that extends landmark-based admissible classical planning heuristics to derive admissible cost estimates for HGN planning problems. To the best of our knowledge, \hgnlmheur\ is the first non-trivial admissible hierarchical planning heuristic.

    \item \textbf{Optimal Planning Algorithm}: We introduce \gdpopt, an A$^*$ search algorithm that uses \hgnlmheur\ to generate \textit{hierarchically-optimal} plans.

    \item \textbf{Experimental Evaluation}: We describe an empirical study on three benchmark planning domains in which \hopgdp\ outperforms optimal classical planners due to its ability to exploit hierarchical knowledge. We also found that \hgnlmheur\ provides useful search guidance; despite substantial computational overhead, it compares favorably in terms of runtime and nodes explored to \gdpblind, using  the trivial heuristic $h=0$.

\end{itemize}

\section{Preliminaries}
In this section we detail the classical planning model, review how landmarks are constructed for classical planning and
an admissible landmark-based heuristic $h_L$, and describe HGN planning using examples from assembly planning.

\subsection{Classical Planning}
We define a {\em classical planning domain} \cpdomain\ as
a finite-state transition system in which each state $s$
is a finite set of ground atoms of a first-order language $L$, and each action $a$ is a ground instance of a planning operator $o$.
A planning operator is a 4-tuple $o = (\head(o), \pre(o), \eff(o), \cost(o))$, where $\pre(o)$ and $\eff(o)$ are conjuncts of literals
called $o$'s {\em preconditions} and {\em effects}, and $\head(o)$ includes $o$'s {\em name} and {\em argument list}
(a list of the variables in $\pre(o)$ and $\eff(o)$). \cost(o) represents the non-negative cost of applying operator $o$.

\mypar{Actions}
An action $a$ is executable in a state $s$ if $s \models \pre(a)$, in which case the resulting state is
$\gamma(a) = (s - \eff^-(a)) \cup \eff^+(a)$, where $\eff^+(a)$ and $\eff^-(a)$ are the atoms and negated atoms, respectively, in $\eff(a)$.
A plan $\pi = \langle a_1, \ldots, a_n\rangle$ is executable in $s$ if each $a_i$ is executable in the state produced by $a_{i-1}$; and in
this case $\gamma(s,\pi)$ is the state produced by executing $\pi$. 
If $\pi$ and $\pi'$ are plans or actions, then their concatenation is $\pi \circ \pi'$.

We define the \textit{cost} of $\pi = \langle a_1, \ldots, a_n\rangle$ as the sum of the costs of the actions in the plan, i.e.
$\cost(\pi) = \sum_{i \in \{1 \ldots n\}} a_i$.

\subsection{Generating Landmarks for Classical Planning}
\label{subsec:lmgen-algs}
There are several landmark generation algorithms suggested in the
literature~\cite{hoffmann04landmarks,richter10lama}.
The general approach used in generating sound landmarks
is to relax the planning problem, generate sound landmarks for the relaxed version, and then use those
for the original planning problem. In this paper, we use LAMA's landmark generation
algorithm~\cite{richter10lama},
which uses relaxed planning graphs and domain-transition graphs in tandem to generate landmarks.

\subsection{\classicallmheur: an Admissible Landmark-based Heuristic for Classical Planning}
We provide some background on $\classicallmheur$, the landmark-based admissible heuristic for classical
planning problems proposed by Karpas and Domshlak~\cite{karpas09costoptimal} that we will be using in our heuristic.

Consider a classical planning problem $P = (\cpdomain,s_0,g)$ and a landmark graph $LG=(L,Ord)$ computed
using any of the off-the-shelf landmark generation algorithms mentioned in Section~\ref{subsec:lmgen-algs}. Then, we can define $\unreachedlm(L,s,\pi) \subseteq L$ to be the set of landmarks
that need to be achieved from $s$ onwards, assuming we got to $s$ using $\pi$.
Note that $\unreachedlm(L,s,\pi)$ is path-dependent:
it can vary for the same state when reached by different paths. It can be computed as follows:
\begin{align*}
\unreachedlm(L,s,\pi) = L &\setminus \\
(\acceptedlm(L,s,\pi) &\setminus \requiredagainlm(L,s,\pi))
\end{align*}

where \acceptedlm$(L,s,\pi) \subseteq L$ is the set of landmarks that were true at some point along $\pi$.
\requiredagainlm$(L,s,\pi) \subseteq L$ is the set of landmarks that were accepted but are required again;
an accepted landmark $l$ is required again if
(1) it does not hold true in $s$, and
(2) it is greedy-necessarily ordered before another landmark $l'$ in $L$ that is not accepted.

Karpas and Domshlak show that
it is possible to partition the costs of the actions $A$ in $\cpdomain$ over the landmarks in $\unreachedlm(L,s,\pi)$ to derive an admissible
cost estimate for the state $s$ as follows:
let $cost(\phi)$ be the cost assigned to the landmark $\phi$, and $cost(a,\phi)$ be the portion of $a$'s cost assigned
to $\phi$. Furthermore, let us suppose these costs satisfy the following set of inequations:
\begin{align}
    \label{eqn:lm-cost-partition}
    \begin{split}
    \forall a \in A : \sum_{\phi \in \unreachedlm(a|L,s,\pi)} cost(a,\phi) \leq &\cost(a) \\
    \forall \phi \in \unreachedlm(L,s,\pi) : cost(\phi) \leq \min_{a \in ach(\phi | s,\pi)} &cost(a,\phi)
    \end{split}
\end{align}

where $ach(\phi | s,\pi) \subseteq A$ is the set of possible achievers of $\phi$ along any suffix of $\pi$, and
$ach(a | L,s,\pi) = \{\phi \in \unreachedlm(L,s,\pi) | a \in ach(\phi|s,\pi)\}$. 

Informally, what these equations are encoding is a scheme to partition the cost of each action across all
the landmarks it could possibly achieve, and assigns to each landmark $\phi$ a cost no more than the minimum cost
assigned to $\phi$ by all its achievers. Given this, they prove the following useful theorem:

\begin{theorem}\label{thm:hl-admissibility}
    Given a set of action-to-landmark and landmark-to-action costs satisfying Eqn.~\ref{eqn:lm-cost-partition},
    $h_{L}(L,s,\pi) = cost(\unreachedlm(L,s,\pi)) = \sum_{\phi \in \unreachedlm(L,s,\pi)} cost(\phi)$ is an admissible estimate of the optimal plan cost from $s$.
\end{theorem}

Note that the choice of exactly how to do the cost-partitioning is left open. One of the schemes Karpas and Domshlak
propose is an \textit{optimal cost-partitioning}
scheme that uses an LP solver to solve the constraints in Eqn.~\ref{eqn:lm-cost-partition} with the objective function
$\max \sum_{\phi \in L(s,\pi)} cost(\phi)$.  This has the useful property that given two sets of landmarks $L$ and $L'$, if
$L \subseteq L'$, then $h_L(L,s,\pi) \leq h_L(L',s,\pi)$. In other words, the more landmarks you provide to $h_L$, the more
informed the heuristic estimate.

\subsection{Goal Networks and HGN Methods}
We extend the definitions of HGN planning~\cite{shivashankar2012hierarchical} to work with
partially-ordered sets of goals, which we call a goal network.

A {\em goal network} is a way to represent the objective of satisfying a partially ordered multiset of goals.
Formally, it is a pair $gn = (T, \prec)$ such that:
\begin{itemize}
	\item $T$ is a finite nonempty set of nodes;
	\item each node $t \in T$ contains a {\em goal} $g_t$ that is a DNF (disjunctive normal form) formula over ground literals;
	\item $\prec$ is a partial order over $T$.
\end{itemize}

\begin{figure}
\centerline{\includegraphics[scale=0.3]{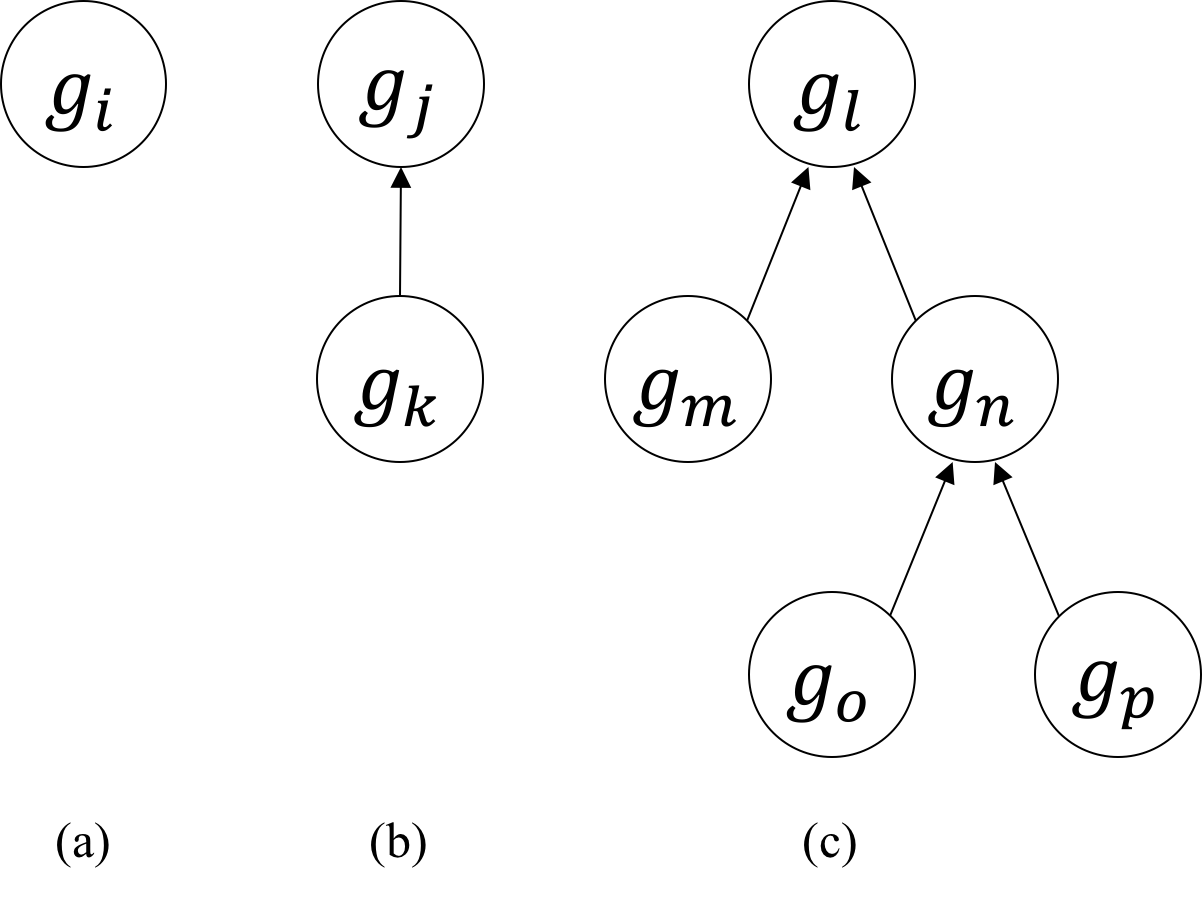}}
\caption{Three generic goal networks we use for examples of the various relationships within a goal network. 
}
\label{fig:hgnexample}
\end{figure}

We will provide examples of both generic and concrete goal networks.
Figure~\ref{fig:hgnexample} shows three generic goal networks.  
Each subfigure is itself a goal network denoted $gn_a, gn_b, gn_c$.
Directed arcs indicate a subgoal pair (e.g., $(g_k, g_j)$ from $gn_b$) such that the first goal must be satisfied before the second goal.
Consider the network $gn_b$ where $g_k$ is a subgoal of $g_j$, then $gn_b = (\{g_j,g_k\}, (g_k \prec g_j))$.  
Network $gn_c$ shows a partial ordering, where $(\{g_m, g_n\} \prec g_l)$.
Similarly,  $(\{g_o, g_p\} \prec g_n)$ and this implies both must occur before $g_l$.
Consider a network $gn_x$ that is composed of $gn_a$ and $gn_b$.
Then $gn_x = (\{g_i,g_j,g_k\}, g_k \prec g_j$).
Note that $gn_x$ is a partially ordered forest of goal networks.

\begin{figure}[!htpb]
    \begin{center}
        \includegraphics[scale=0.38]{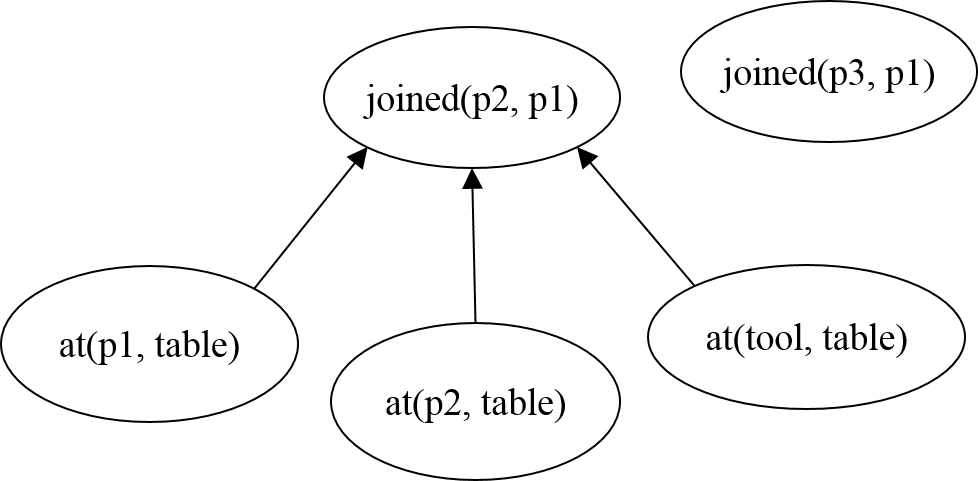}
    \end{center}
    \caption{Goal Network for an Automated Manufacturing domain}
    \label{fig:sample-gn}
\end{figure}
Figure~\ref{fig:sample-gn} shows a concrete goal network for an automated manufacturing domain.
\textsf{joined}$(x,y)$ denotes the goal of assembling the parts $x$ and $y$ together, while \textsf{at}$(x,loc)$
represents the goal of getting $x$ to location $loc$.
In this goal network, \textsf{joined}$(p_2,p_1)$ and \textsf{joined}$(p_3,p_1)$ are unordered
with respect to one another. Furthermore, \textsf{joined}$(p_2,p_1)$ has three subgoals that need to be achieved
before achieving it, i.e the goals of getting the parts $p_1$, $p_2$ and the $tool$
to the assembly table. These subgoals are also unordered with respect to one another, indicating that
the goals can be accomplished in any order.

\paragraph{HGN Methods}
An {\em HGN method} $m$ is a 4-tuple $(\head(m),$ $\goal(m),$ $\pre(m),$ $\network(m))$
where the head $\head(m)$ and preconditions $\pre(m)$ are similar to those of a planning operator. $\goal(m)$ 
is a conjunct of literals representing the goal $m$ decomposes. $\network(m)$ is the goal network that $m$
decomposes into. 
By convention, $\network(m)$ has a last node $t_g$ containing the goal $\goal(m)$ to ensure that $m$ accomplishes its own goal.

\begin{figure}[!htpb]
    \begin{center}
        \includegraphics[width=\linewidth]{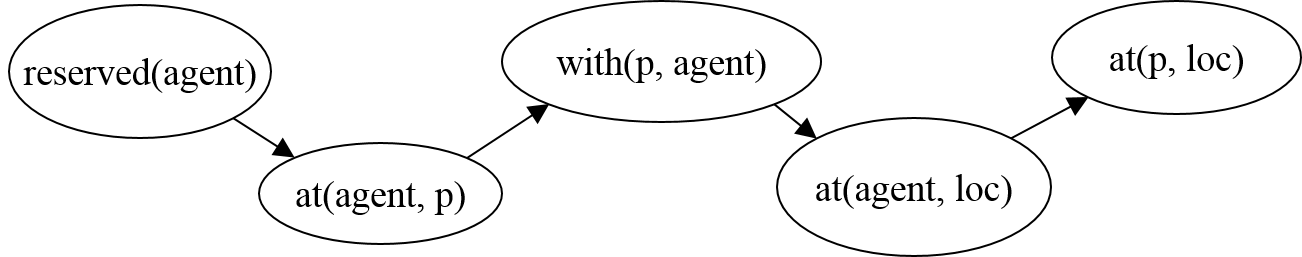}
    \end{center}
    \caption{Subgoal network of \textsf{deliver-obj}$(p,loc,agent)$, an HGN method to deliver the part
    $p$ to $loc$ using $agent$.}
    \label{fig:sample-method}
\end{figure}

Figure~\ref{fig:sample-method} describes the goal network that the \textsf{deliver-obj} method,
a method responsible for solving problems related to delivering parts and tools to their destinations,
decomposes a goal into. 
This method is relevant to \textsf{at}$(x,loc)$ goals (since that's the last node), and
its preconditions are
$\pre(\textsf{deliver-obj}) = \{\neg\textsf{reserved}(agent),\textsf{can-carry}(agent,p) \ldots\}$.  


Whether a node has predecessors impacts the kinds of operations we allow.  
We refer to any node in a goal network $gn$ having no predecessors as an \textit{unconstrained} node of $gn$, otherwise the node is \textit{constrained}.
The constrained nodes of  Figure~\ref{fig:hgnexample}  include $g_j, g_l, g_n$ and the remaining are unconstrained.
The unconstrained nodes in Figure~\ref{fig:sample-gn} include all the \textsf{at} nodes
as well as the \textsf{joined}$(p3,p1)$ node.

We define the following operations over any goal network $gn=(T,\prec)$:
\begin{enumerate}
    \item {\bf Goal Release}: Let $t \in T$ be an unconstrained node. Then the removal of $t$ from $gn$, denoted by $gn-t$,
        results in the goal network $gn' = (T',\prec')$ where $T' = T\setminus\{t\}$ and $\prec'$ is the restriction of $\prec$ to $T'$.
    \item {\bf Method Application}: Let $t \in T$ be an unconstrained node.
        Also, let $m$ be a method applied to $t$ with 
        $\network(m) = (T_m,\prec_m)$. Finally,
        recall that $\network(m)$ always contains a 'last' node that contains $\goal(m)$;
        let $t_g$ be this node.
        Then the application of $m$ to $gn$ via $t$, denoted by $gn\ \circ_t\ m$,
        results in the goal network $gn' = (T', \prec')$ where $T' = T \cup T_m$
        and $\prec' = \prec \cup \prec_m \cup \{(t_g, t)\}$.
        Informally, this operation adds the elements of $\network(m)$ to $gn$, preserving the
        order specified by $\sub(m)$ and setting $\goal(m)$ as a predecessor of $t$. 
\end{enumerate}

\subsection{HGN Domains, Problems and Solutions}
A {\em HGN domain} is a pair $D=(\cpdomain,M)$ where
\cpdomain\ is a classical planning domain and $M$ is a set of HGN methods. 

A {\em HGN planning problem} is a triple $P = (D,s_0,gn_0)$, where $D$ is an HGN domain, $s_0$ is the initial state,
and $gn_0=(T,\prec)$ is the initial goal network.
\begin{definition}[Solutions to HGN Planning Problems]
	\label{dfn:hgn-solns}
The set of {\em solutions} for $P$ is defined as follows:
\begin{description}
	\item[Base Case.] If $T$ is empty, the empty plan is a solution for $P$.
\end{description}
In the following cases, let $t \in T$ be an unconstrained node.
\begin{description}
	\item[Unconstrained Goal Satisfaction.] If $s_0 \models g_t$, then any solution for $P'=(D, s_0, gn_0-t)$ is also a solution for $P$.	
    \item[Action Application.] 
        If action $a$ is applicable in $s_0$ and $a$ is relevant to $g_t$, and $\pi$ is a solution for 
        $P'=(D, \gamma(s_0,a), gn_0)$, then $a\circ \pi$ is a solution for $P$.
    \item[Method Decomposition.] If $m$ is a method applicable in $s$ and relevant to $g_t$, then any solution to 
        $P' = (D, s_0, gn_0 \circ_t m)$ is also a solution to $P$.
\end{description}
\end{definition}


Note that HGN planning allows an action to be applied only if it is \textit{relevant} to an
unconstrained node in $gn$;
this prevents unrestricted chaining of applicable actions as done in classical planning
and allows for tighter control of solutions as in HTN planning.

Let us denote $\sol(P)$ as the set of solutions to an HGN planning problem $P$
as allowed by Definition~\ref{dfn:hgn-solns}.
Then we can define what it means for a solution $\pi$
to be \textit{hierarchically optimal} with respect to $P$ as follows:

\begin{definition}[Hierarchically Optimal Solutions]
    A solution $\pi^{h,*}$ is hierarchically optimal with respect to $P$ if 
    $\pi^{h,*} = \text{argmin}_{\pi \in \sol(P)} \cost(\pi)$.
\end{definition}

\section{The \gdpopt\ Algorithm}
\label{sec:hopgdp}

Algorithm~\ref{alg:gdpopt} describes \gdpopt. It takes as input an HGN domain $D=(D',M)$,
the initial state $s_0$ and the initial goal network $gn_0$.
It does an A$^*$ search using the admissible HGN heuristic \hgnlmheur\ (described in Section~\ref{sec:hHL})
to compute a hierachically optimal solution to the problem; it either
returns a plan if it finds one,
or \failure\ if the problem is unsolvable.

\mypar{Initialization}
It starts off by initializing \openlist\ (Line~\ref{alg:gdpopt:init-open}), which is a priority queue that sorts the HGN search nodes
yet to be expanded by their $f$-value, where $f((s,gn,\pi)) = \cost(\pi) + \hgnlmheur(s,gn)$.
\openlist\ initially contains the initial search node $(s_0,gn_0,\langle\rangle)$.
It also initializes \searchSpace\ (Line~\ref{alg:gdpopt:init-ss}), the set of all nodes seen during the search process.
This data structure keeps track of the best known path for each $(state,goal\text{-}network)$ pair,
and is thus helpful to detect when we find a cheaper path to a previously seen HGN search node.

\begin{algorithm}[!htbp]
    \caption{Pseudocode of \gdpopt. It takes as arguments the domain description
    $D=(\cpdomain,M)$, the initial state $s_0$, and the initial goal network $gn_0$. It either returns
    a plan if it finds one, or \failure\ if it doesn't.}
    \label{alg:gdpopt}
    \begin{algorithmic}[1]
        \Function{\gdpopt}{$D, s_0, gn_0$}
        \State $\openlist \leftarrow (s_0,gn_0,\langle\rangle)$\label{alg:gdpopt:init-open}
        \State $\searchSpace \leftarrow (s_0,gn_0,\langle\rangle)$\label{alg:gdpopt:init-ss}

        \While{$\openlist$ is not empty}\label{alg:gdpopt:iter-start}
        \State rem. $(s,gn,\pi)$ with lowest $f$-value from \openlist\label{alg:gdpopt:rem-best-node}

            \If{$gn$ is empty}\label{alg:gdpopt:base-case}
            \ \Return $\pi$ 
            \EndIf

            \State successors $\leftarrow \getSuccessors(D,s,gn,\pi)$\label{alg:gdpopt:getsucc}
            \For{$(s',gn',\pi') \in$ successors}
                \If{$\exists (s',gn',\eta) \in \searchSpace$}\label{alg:gdpopt:new-node-check}
                    \If{$\cost(\pi') < \cost(\eta)$}\label{alg:gdpopt:cost-check}
                        \State replace $(s',gn',\eta)$ with $(s',gn',\pi')$ 
                        \text{\ \ \ \ \ \ \ \ \ \ \ \ \  \ \ \ \ \ \ \ \ \ \ \ \ \ \ \ \ } in \searchSpace
                    \Else\ \textbf{continue}\label{alg:gdpopt:new-path-costlier}
                    \EndIf
                \Else
                \ add $(s',gn',\pi')$ to \searchSpace\label{alg:gdpopt:add-new-to-ss}
                \EndIf

                \State eval. $f$-value of $(s',gn',\pi')$ and add to \openlist\label{alg:gdpopt:add-node-to-open}
            \EndFor
        \EndWhile\label{alg:gdpopt:iter-end}

        \State \Return \failure\label{alg:gdpopt:fail}

        \EndFunction \\

        \Function{\getSuccessors}{$D,s,gn,\pi$}
            \State successors $\leftarrow \emptyset$

            \For{unconstrained $g \in gn$ satisfied in $s$}\label{alg:getsucc:goal-satisfaction-start}
                \State add the node $(s,gn-\{g\}, \pi)$ to successors
            \EndFor\label{alg:getsucc:goal-satisfaction-end}

            \State $\mathcal{A} \leftarrow$ actions in $D$ applicable in $s$ and relevant to an
            unconstrained goal in $gn$\label{alg:getsucc:compute-actions}
            \For{$a \in \mathcal{A}$}\label{alg:getsucc:action-application-start}
                \State add the node $(\gamma(s,a),gn,\pi \circ a)$ to successors
            \EndFor\label{alg:getsucc:action-application-end}

            \State $\mathcal{M} \leftarrow\{ (m,g)$ s.t. $m \in M$ is applicable in $s$ and relevant to an
            unconstrained goal $g$ in $gn\}$\label{alg:getsucc:compute-methods}
            \For{$(m,g) \in \mathcal{M}$}\label{alg:getsucc:goal-decomposition-start}
                \State add the node $(s, gn \circ_g m, \pi)$ to successors
            \EndFor\label{alg:getsucc:goal-decomposition-end}
            \State \Return successors\label{alg:getsucc:return}
        \EndFunction
    \end{algorithmic}
\end{algorithm}

\mypar{Search}
\gdpopt\ now proceeds to do an A$^*$ search in the space of HGN search nodes starting
from the initial node. While \openlist\ is not empty, it does the following
(Lines~\ref{alg:gdpopt:iter-start}--\ref{alg:gdpopt:iter-end}): it removes the HGN search node $N = (s,gn,\pi)$
with the best $f$-value from \openlist\ (Line~\ref{alg:gdpopt:rem-best-node}) and first checks
if $gn$ is empty (Line~\ref{alg:gdpopt:base-case}). If this is true, this means that all the goals in
$gn_0$ have been solved, and $\pi$ is the optimal solution to the HGN planning problem.

If $gn$ is not empty, then the algorithm proceeds by using
the \getSuccessors\ subroutine to compute $N$'s successor nodes (Line~\ref{alg:gdpopt:getsucc}).
For each successor node $(s',gn',\pi')$, it proceeds to do the following: it checks to see if another path
$\eta$ to $(s',gn')$ exists in \searchSpace\ (Line~\ref{alg:gdpopt:new-node-check}).
If this is the case and
if $\eta$ is costlier than $\pi'$ (Line~\ref{alg:gdpopt:cost-check}), it updates \searchSpace\ with the new path;
and reopens the search node (Line~\ref{alg:gdpopt:add-node-to-open});
if $\eta$ is cheaper than the new plan $\pi'$, it simply skips this successor
(Line~\ref{alg:gdpopt:new-path-costlier}).

If $(s',gn')$ has not been seen before, it adds $N' = (s',gn',\pi')$ to \searchSpace\ to track the currently
best-known plan $\pi'$ to $(s',gn')$
(Line~\ref{alg:gdpopt:add-new-to-ss}). It also evaluates the $f$-value of $N'$
(note that this is where \hgnlmheur\ is called) and
adds it to \openlist\ (Line~\ref{alg:gdpopt:add-node-to-open}).

If there are no more nodes left in \openlist, this implies that it has exhausted the search space without
finding a solution, and therefore returns \failure\ (Line~\ref{alg:gdpopt:fail}).

\mypar{Computing Successors}
The procedure \getSuccessors\ computes the successors of a given HGN search node $(s,gn,\pi)$
in accordance with Definition~\ref{dfn:hgn-solns}.
First, we check to see if there are any unconstrained goals $g$ in $gn$ that are satisfied in the current state $s$.
We then proceed to create new HGN search nodes by removing all such goals from $gn$
(Line~\ref{alg:getsucc:goal-satisfaction-start}--\ref{alg:getsucc:goal-satisfaction-end}).
Next, we compute all actions applicable in $s$ and relevant to an unconstrained goal in $gn$
(Line~\ref{alg:getsucc:compute-actions})
and create new
search nodes by progressing $s$ using these actions
(Line~\ref{alg:getsucc:action-application-start}--\ref{alg:getsucc:action-application-end}).
We compute all pairs $(m,g)$ such that $m$ is an HGN method applicable in $s$ and relevant
to an unconstrained goal $g$ in $gn$
(Line~\ref{alg:getsucc:compute-methods})
and create new
search nodes by decomposing $g$ in $gn$ using $m$ 
(Line~\ref{alg:getsucc:goal-decomposition-start}--\ref{alg:getsucc:goal-decomposition-end}).
Finally, we return the set of generated successor nodes (Line~\ref{alg:getsucc:return}).

\section{\hgnlmheur: An Admissible Heuristic for HGN Planning}
\label{sec:hHL}
\begin{algorithm}
    \caption{Procedure for computing landmarks for relaxed HGN planning problems.}
    \begin{algorithmic}[1]
        \Function{\hgnlmgen}{$s,gn$}
            \State \texttt{queueSeeds} $\leftarrow gn$ \label{alg:clfhp:init-queueseeds}
            \State \texttt{queue} $\leftarrow \emptyset$

            \While{\texttt{queueSeeds} is not empty}\label{alg:clfhp:queueseeds-iter}
                \State choose a $g$ w/o successors from \texttt{queueSeeds}, and
                remove it along with all associated orderings\label{alg:clfhp:rem-queueseeds}
                \State \addlm$(g)$,
                add $g$ to \texttt{queue}\label{alg:clfhp:add-queue}
                \State add any orderings $g$ shares with other goals from $gn$ already
                added to LG
                \While{\texttt{queue} is not empty}\label{alg:clfhp:iter-queue-start}
                    \State pop landmark $\psi$ from \texttt{queue} and use
                    \lmgenclassical\ to generate
                    the new set of landmarks $\Phi$
                    \For{$\phi \in \Phi$}
                    \textsc{addLM}$(\phi, \phi \rightarrow_{gn} \psi)$
                    \EndFor
                    \EndWhile\label{alg:clfhp:iter-queue-end}
            \EndWhile
            \State \Return LG
        \EndFunction \\

        \Function{\addlm}{$\phi$}
        \If{$\phi$ is a fact and $\exists \phi'\in LG: \phi' \neq \phi \land \phi \models \phi'$}
        \label{alg:alm:new-lm-sub-old-start}
            \State remove $\phi'$ from LG and all orderings it is part of\label{alg:alm:new-lm-sub-old-end}
        \EndIf
        \If{$\exists \phi' \in LG: \phi' \models \phi$}\label{alg:alm:old-lm-sub-new-start}
            \Return $\phi'$
        \EndIf\label{alg:alm:old-lm-sub-new-end}
        \If{$\phi \notin LG$}\label{alg:alm:new-lm-start}
            add $\phi$ to \texttt{queue}
            and \Return $\phi$
        \EndIf\label{alg:alm:old-lm-sub-new-end}
        \EndFunction \\

        \Function{\addlmandorder}{$\phi, \phi \rightarrow_{x} \psi$}
            \State $\eta \leftarrow \addlm(\phi)$
            \State add ordering $\eta \rightarrow_x \psi$ to LG
       \EndFunction
    \end{algorithmic}
    \label{alg:hgn-lm-discovery}
\end{algorithm}

As mentioned in Section~\ref{sec:hopgdp}, \hopgdp\ uses \hgnlmheur\ to compute the $h$-values
(and thus, the $f$-values) of search nodes. 
In this section, We will proceed to describe how to construct \hgnlmheur\ as follows:
\begin{enumerate}
    \item 
We define a relaxation of HGN planning that ignores the provided methods and allows
unrestricted action chaining as in classical planning, which expands the set of allowed
solutions,
\item 
    We will extend landmark generation algorithms for classical planning problems to
compute sound landmark graphs for the relaxed HGN planning problems, which in turn are
sound with respect to the original HGN planning problems as well, and finally
\item
We will use admissible classical planning heuristics like $h_L$ on these landmark graphs
to compute admissible cost estimates for HGN planning problems.
\end{enumerate}

\subsection{Relaxed HGN Planning}
\begin{definition}[Relaxed HGN Planning]
    A relaxed HGN planning problem is a triple $P=(\cpdomain,s_0,gn_0)$ where $D$ is a classical planning domain,
    $s_0$ is the initial state, and $gn_0$ is the initial goal network. Any sequence of actions $\pi$
    that is executable in state $s_0$ and achieves the goals in $gn_0$ in an order consistent with
    the constraints in $gn_0$ is a valid solution to $P$.
\end{definition}

Relaxed HGN planning
can thus be viewed as an extension of classical planning to solve for goal networks,
where there are no
HGN methods and the objective is to generate
sequences of actions that satisfy the goals in $gn_0$ in an order consistent with $gn_0$.
In fact, it is easy to show that relaxed HGN planning, in contrast to HGN planning,
is no more expressive than classical planning, and relaxed HGN planning
problems can be compiled into classical planning problems quite easily.

Next, we will show how to leverage landmark generation algorithms for classical planning to generate
landmark graphs for relaxed HGN planning.

\subsection{Generating Landmarks for Relaxed HGN Planning}
\label{sec:hgnlmgen}

This section describes a landmark discovery technique that can use any landmark discovery technique for
classical planning (referred to as \lmgenclassical\ here) such as~\cite{richter10lama}
to compute landmarks for relaxed HGN planning problems. The main difference here is that while classical planning problems
are $(state, goal)$ pairs, relaxed HGN planning problems are $(state, goal\text{-}network)$ pairs;
every goal in the goal network can be thought of as a landmark. Therefore, there is now
a partially ordered set of goals to compute landmarks from, as opposed to a single goal in classical planning.

We therefore need to generalize classical planning landmark generation techniques to work
for relaxed HGN planning problems.
The \hgnlmgen\ algorithm (Algorithm~\ref{alg:hgn-lm-discovery})
describes one such generalization.
At a high level, \hgnlmgen\ proceeds by computing landmark
graphs for each goal $g$ in $gn$ (which in fact is a classical planning problem)
and merging them all together to create the final landmark
graph $LG$.

\hgnlmgen\ takes as input a relaxed HGN planning problem
$(s,gn)$ and generates $LG$, a graph of landmarks.
First,
\texttt{queueSeeds} is initialized with a copy of $gn$ (Line \ref{alg:clfhp:init-queueseeds}).
This is because unlike in classical planning where we
generate landmarks for a
single goal,
in HGN planning we have a partially ordered set of goals to seed
landmark generation; \texttt{queueSeeds} stores these seeds.
We also initialize \texttt{queue}, the openlist of landmarks, to $\emptyset$.

While there is a goal $g$ from $gn$ that we have not yet computed landmarks for
(Line~\ref{alg:clfhp:queueseeds-iter}), we do the following: we remove it from
\texttt{queueSeeds} along with all induced orderings and add it to \texttt{queue}
(Lines~\ref{alg:clfhp:rem-queueseeds}--\ref{alg:clfhp:add-queue}).
We also add $g$ to $LG$ using \addlm; we also add any ordering constraints it might have with other
elements of $gn$ that have already been added to $LG$.
This \texttt{queue}
is then used as a starting point by \lmgenclassical\ to begin landmark generation. We iteratively
use \lmgenclassical\ to pop landmarks off the \texttt{queue} and generate new landmarks by backchaining until
we can no longer generate any more landmarks
(Lines~\ref{alg:clfhp:iter-queue-start}--\ref{alg:clfhp:iter-queue-end}).
Each new landmark is added to $LG$ by the \addlmandorder\ procedure. Once all goals
in $gn$ have been handled, the landmark generation process is completed and the algorithm
returns $LG$.

The \addlm\ procedure takes as input a computed
landmark $\phi$, adds it to $LG$ and returns a landmark $\eta$.
There are three cases to consider:
\begin{itemize}
    \item $\phi$ subsumes another landmark $\phi'$ in $LG$, implying we can
        remove $\phi'$ and replace it with $\phi$ (since $\phi$ is a stronger version of $\phi'$),
        and return $\phi$
        (Lines~\ref{alg:alm:new-lm-sub-old-start}--\ref{alg:alm:new-lm-sub-old-end})
    \item $\phi$ is subsumed by another landmark $\phi'$ in $LG$, implying we can
        ignore $\phi$ (Lines~\ref{alg:alm:old-lm-sub-new-start}).
        In this case, we don't add any new landmark to $LG$ and simply return $\phi'$
    \item $\phi$ is a new landmark, in which case we can simply add it to $LG$ and return $\phi$
        (Lines~\ref{alg:alm:new-lm-start})
\end{itemize}

The \addlmandorder\ procedure takes as input a landmark $\phi$ and an ordering constraint
$\phi \rightarrow_x \psi$ and adds them to $LG$. More precisely, it adds $\phi$ to $LG$ using
\addlm, which returns the added landmark $\eta$. It then adds the ordering constraint between
$\eta$ and $\psi$ in $LG$.

\begin{figure}[!htpb]
    \centering
\subfloat[]{
            \label{fig:lm-graph-toplevel-goal}
            \includegraphics[scale=0.32]{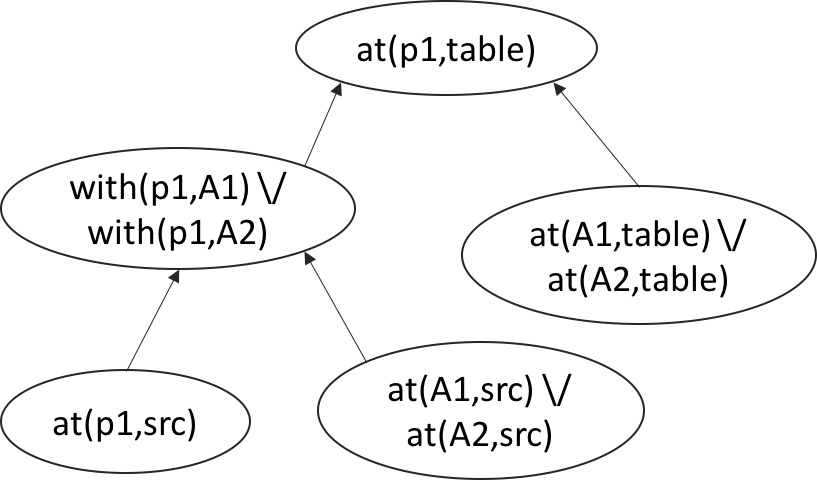}}
\subfloat[]{
            \label{fig:lm-graph-after-method}
        \includegraphics[scale=0.32]{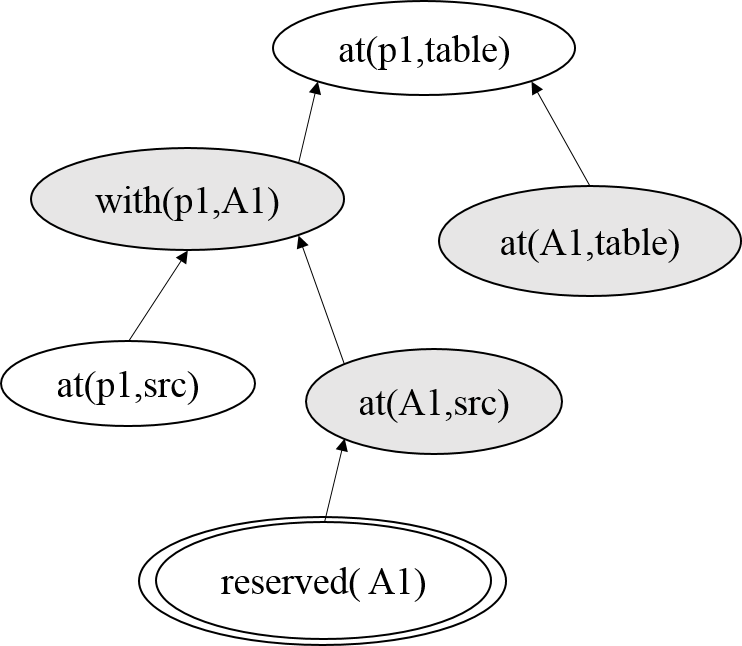}}
    \caption{(a) LM graph on goal network containing a single goal \textsf{at}$(p_1,table)$.
             (b) LM graph after decomposing \textsf{at}$(p_1,table)$ with 
             \textsf{deliver-obj}$(p_1,table,A1)$.
The double-circled landmarks represent new landmarks inferred after the method decomposition,
while the landmarks colored gray are new landmarks
that subsumed an existing one in (a).}
    \label{fig:lm-graph}
\end{figure}

\mypar{LM graph computation example}
Figure~\ref{fig:lm-graph} illustrates the working of \hgnlmgen. Let us assume the goal network $gn$
contains only one goal $g=\textsf{at}(p_1,table)$.
Figure~\ref{fig:lm-graph-toplevel-goal} illustrates the output of \hgnlmgen\ on $g$. This is identical to what
\lmgenclassical\ would generate, since $gn$ contains only one goal, making the relaxed HGN problem equivalent
to a classical planning problem.

Now, let us assume that we decompose $gn$ using the $m=\textsf{deliver-obj}(p_1,table,A1)$, and get the new
goal network $gn'$, which essentially looks like an instantiated version of the network
in Figure~\ref{fig:sample-method}. Now if we run \hgnlmgen\ on $gn'$, we end up generating the landmark
graph in Figure~\ref{fig:lm-graph-after-method}, which is a more focused version of the first landmark graph.
This is because the goals in $gn'$ are landmarks that must be accomplished, which constrains the set of valid
solutions that can be generated. For instance, since we've committed to agent $A1$, every solution we can generate
from $gn'$ will involve the use of $A1$. We can, as a result, generate more focused landmarks than we otherwise could
have from just the top-level goal $g$. This includes fact landmarks that replace disjunctive landmarks
(the ones in gray in Fig.~\ref{fig:lm-graph-after-method}) as well as completely new landmarks that arise
as a result of the method; e.g. \textsf{reserved}$(A1)$ is not a valid landmark for $gn$, but is one
for $gn'$.

An important point to note at this point is that the subgoals in $gn'$ are not \textit{true} landmarks for $g$;
they are landmarks once we commit to applying method $m$. However, this actually ends up being useful to us, since
it allows us to generate different landmark graphs for different methods; for instance, if we had committed to $A2$,
we would have obtained a different set of landmarks specific to $A2$. Now, landmark-based heuristics when applied to these
two graphs would get us different heuristic estimates, thus allowing to differentiate between these two methods
by using the specific subgoals each method introduces.

It is easy to show that \hgnlmgen\ generates sound landmark graphs
for relaxed HGN planning problems:
\begin{claim}
    \label{thm:relaxed-hgn-lmgraph-soundness}
    Given a relaxed HGN planning problem $P=(\cpdomain,s_0,gn_0)$, $LG=\hgnlmgen(s_0,gn_0)$ is a sound
    landmark graph
    for $P$.
\end{claim}

Let $P=((\cpdomain,M), s_0, gn_0)$ be an HGN planning problem, and let $P'=(\cpdomain,s_0,gn_0)$ be
the corresponding
relaxed version. Then by definition, any solution to $P$ is a solution to $P'$. Therefore, it
is easy to see that a landmark of $P'$ is also a sound landmark of $P$.
More generally, a landmark graph generated for $P'$ is going to be sound with respect to $P$ as well:
\begin{claim}
    \label{thm:hgn-lmgraph-soundness}
    Given an HGN planning problem $P$, then $LG=\hgnlmgen(s_0,gn_0)$ is a sound
    landmark graph for $P$.
\end{claim}

\subsection{Computing \hgnlmheur}
\label{sec:hgnlmheur}
The main insight behind \hgnlmheur\ is the following: 
\textit{since the \hgnlmgen\ algorithm generates sound landmarks and orderings for relaxed
(and therefore regular) HGN planning problems, we can use any
admissible landmark-based heuristic from classical planning to derive an admissible cost estimate for
HGN planning problems.}

In particular, \hgnlmheur\ uses \classicallmheur\ as follows:
given an HGN search node $(s,gn)$, the landmark graph is given by
$LG_{HGN}=\hgnlmgen(s,gn)$. Then
\begin{align}\label{dfn:hgnlmheur}
\hgnlmheur(s,gn,\pi) = h_L(LG_{HGN},s,\pi)
\end{align}
where $\pi$ is the plan generated to get to $(s,gn)$.

A couple of important implementation details:
when using \classicallmheur\ to guide classical planners, it is sufficient to compute
the landmark graph just once upfront since it can be reused in every state along the plan
due to the goal staying the same. This isn't the case in HGN planning; method decomposition
can change the goal network. So, \hgnlmheur\ requires re-computing the landmark graph
each node. In our implementation, we try to optimize this process by computing landmark graphs
for each goal network we encounter from the initial state and caching them for use in future nodes
containing the same goal network. Section~\ref{subsec:expt-takeaways} discusses the impact of this
overhead in the experiments. Secondly, while the optimal cost partitioning scheme in
\classicallmheur\ provides more informed
heuristic estimates, we chose to use the uniform cost partitioning scheme in our
implementation since the former requires solving an LP at each search node, which is
costly.

\subsection{Admissibility of \hgnlmheur}
Claim~\ref{thm:hgn-lmgraph-soundness} shows that given an HGN problem
$P=(D,s_0,gn_0)$, $LG=\hgnlmgen(s_0,gn_0)$ is a sound landmark graph with respect to $P$. Furthermore,
Lemma~\ref{thm:hl-admissibility} shows that $h_L(LG,s_0,\langle\rangle)$ provides an admissible
cost estimate of the optimal plan starting from $s_0$ that achieves all the landmarks in $LG$.
Since every solution to $P$ has to achieve all the landmarks in $LG$ in a consistent order,
$h_L(LG,s_0,\langle\rangle)$ provides an admissible estimate of the optimal cost to $P$ as well.
However, from Eq.~\ref{dfn:hgnlmheur}, $h_L(LG,s_0,\langle\rangle) = \hgnlmheur(s_0,gn_0,\langle\rangle)$.
Therefore, we
have the following theorem:
\begin{theorem}[Admissibility of \hgnlmheur]
    Given an HGN planning domain $D$, a search node $(s,gn,\pi)$ and its cost-optimal solution $\pi^{*,HGN}_{s,gn}$,
    $\hgnlmheur(s,gn,\pi) \leq \pi^{*,HGN}_{s,gn}$.
\end{theorem}

\begin{figure*}[!htpb]
    \centering
\subfloat[]{
            \label{fig:logistics-nodes-expanded}
            \includegraphics[scale=0.40]{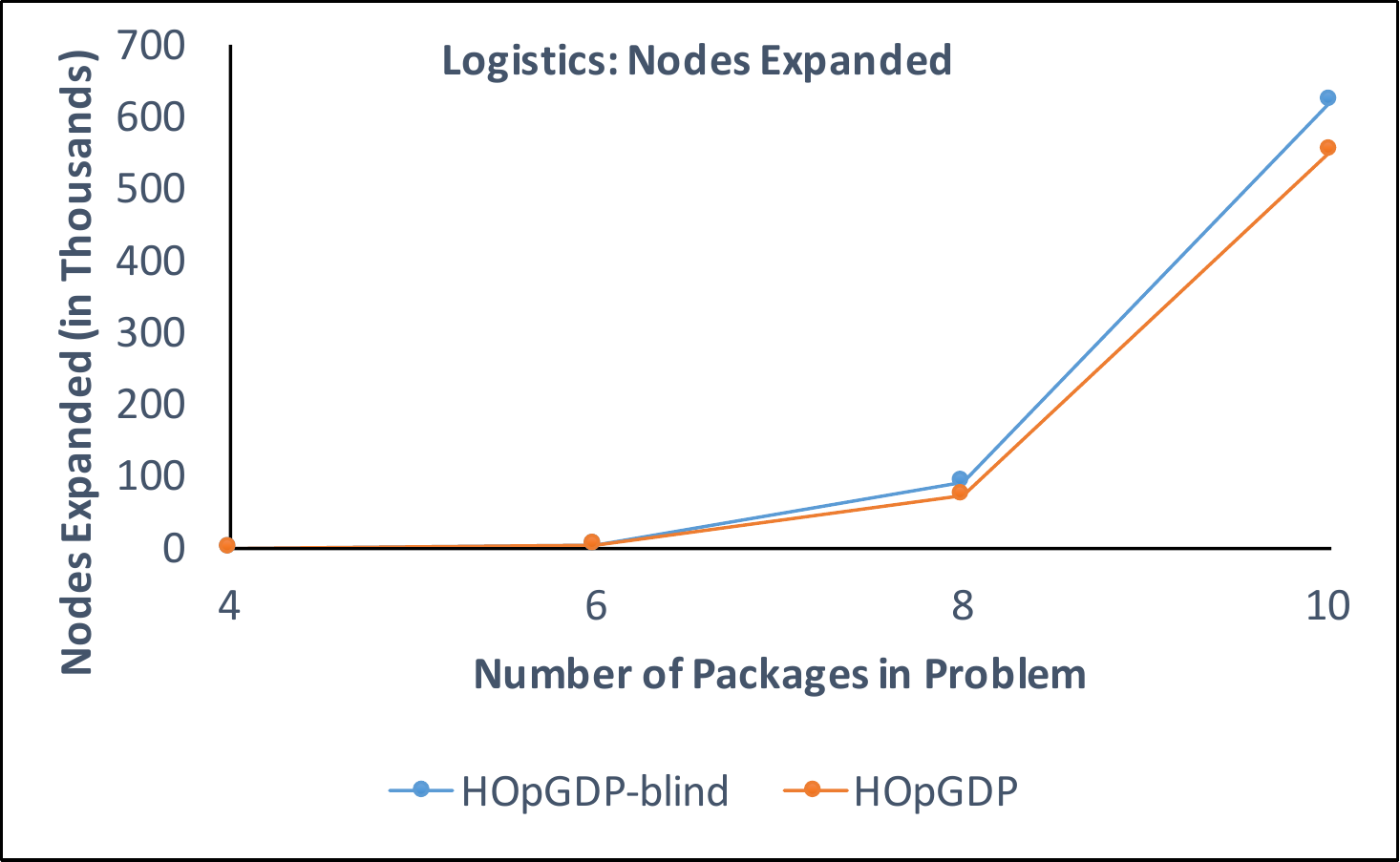}}
\subfloat[]{
            \label{fig:bw-nodes-expanded}
            \includegraphics[scale=0.40]{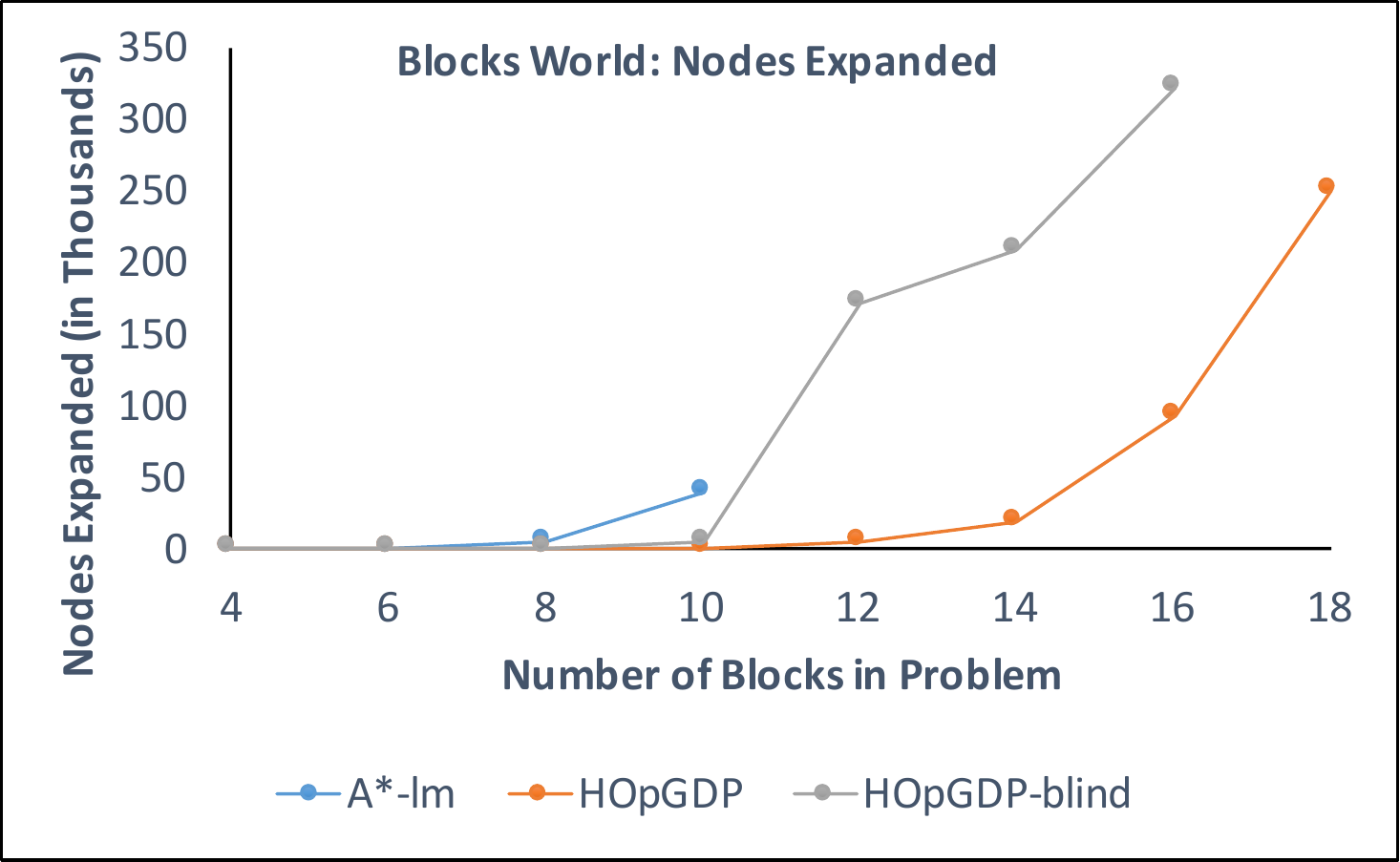}}
\subfloat[]{
            \label{fig:depots-nodes-expanded}
            \includegraphics[scale=0.40]{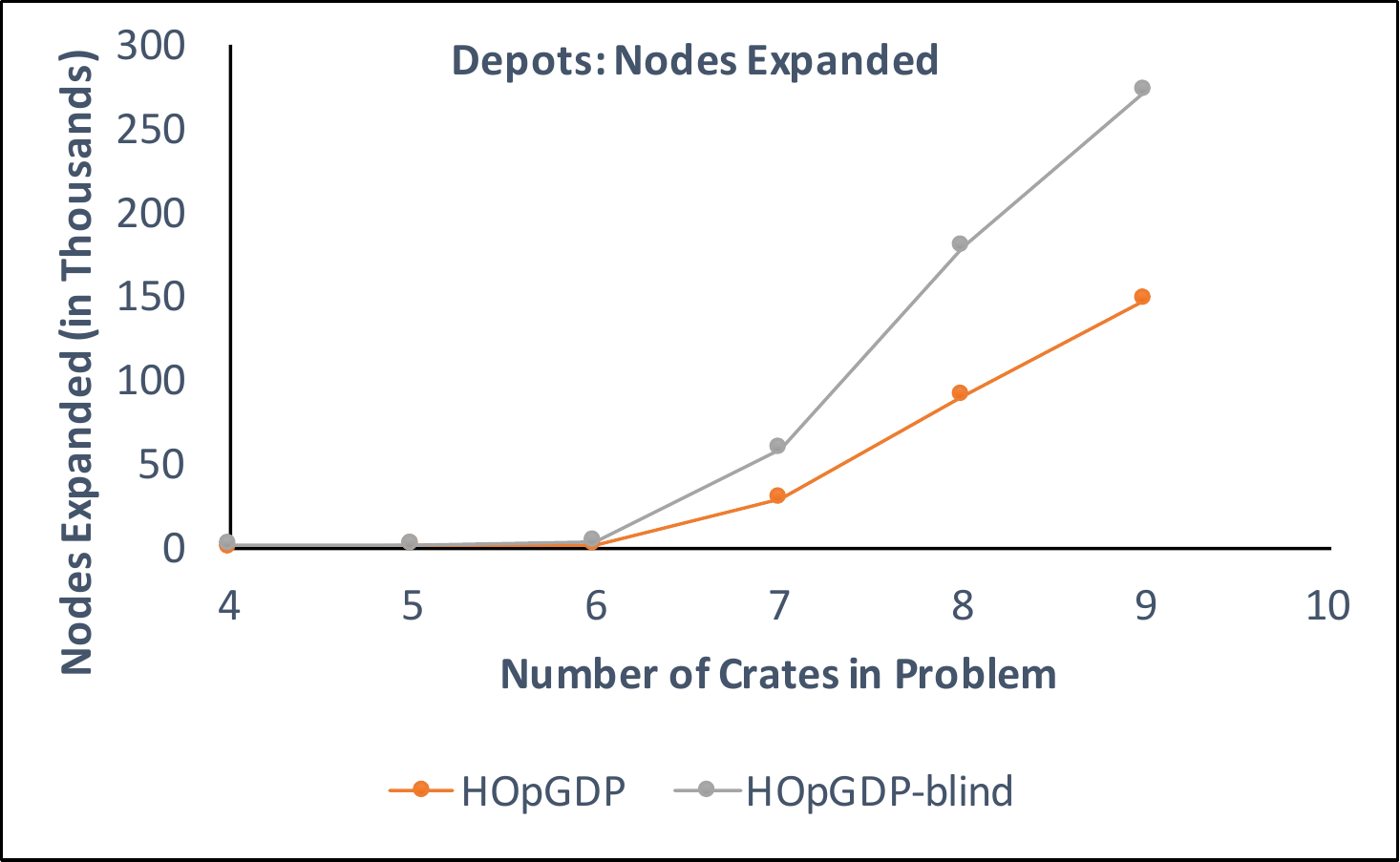}}

\subfloat[]{
            \label{fig:logistics-running-time}
            \includegraphics[scale=0.40]{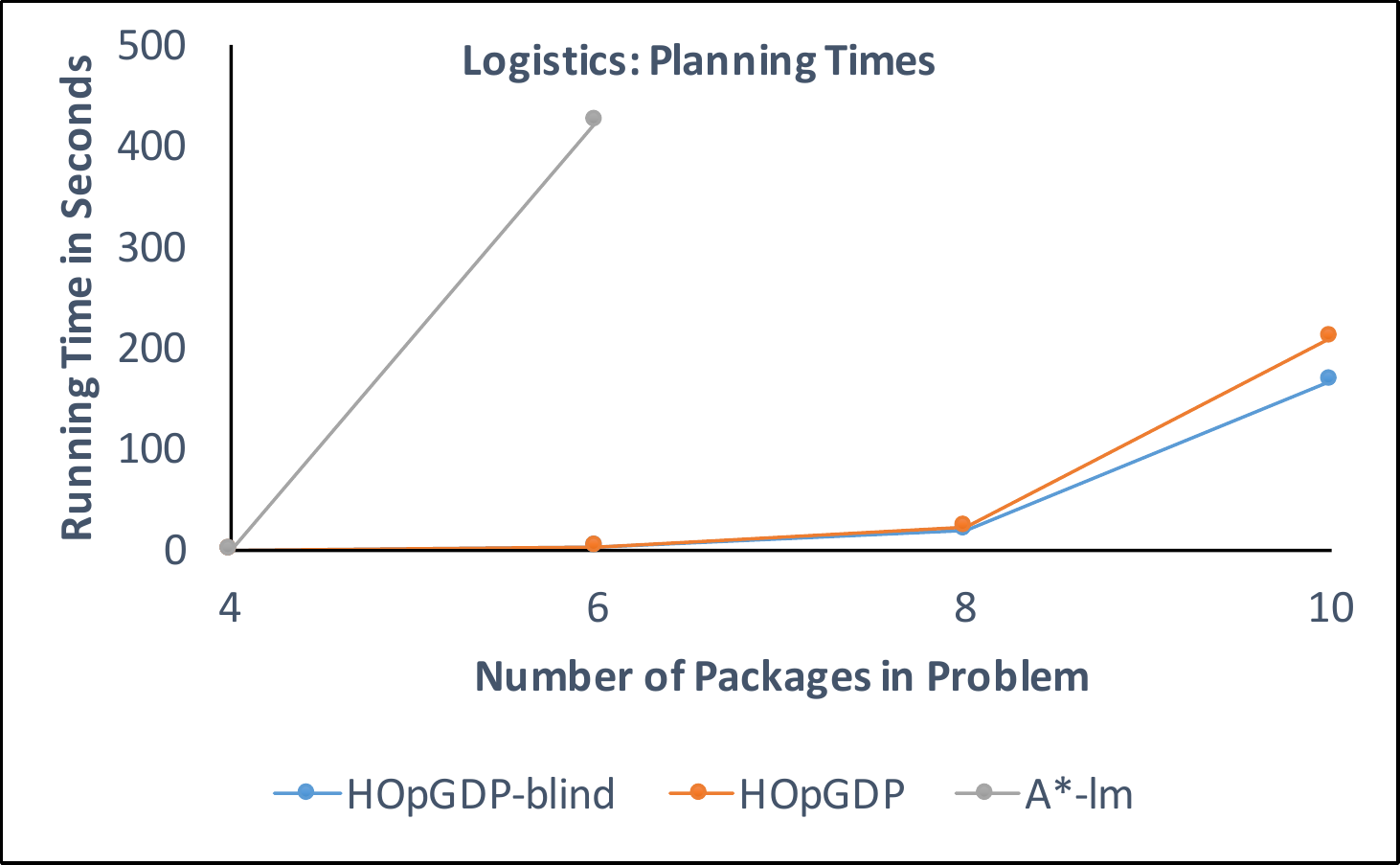}}
\subfloat[]{
            \label{fig:bw-running-time}
            \includegraphics[scale=0.40]{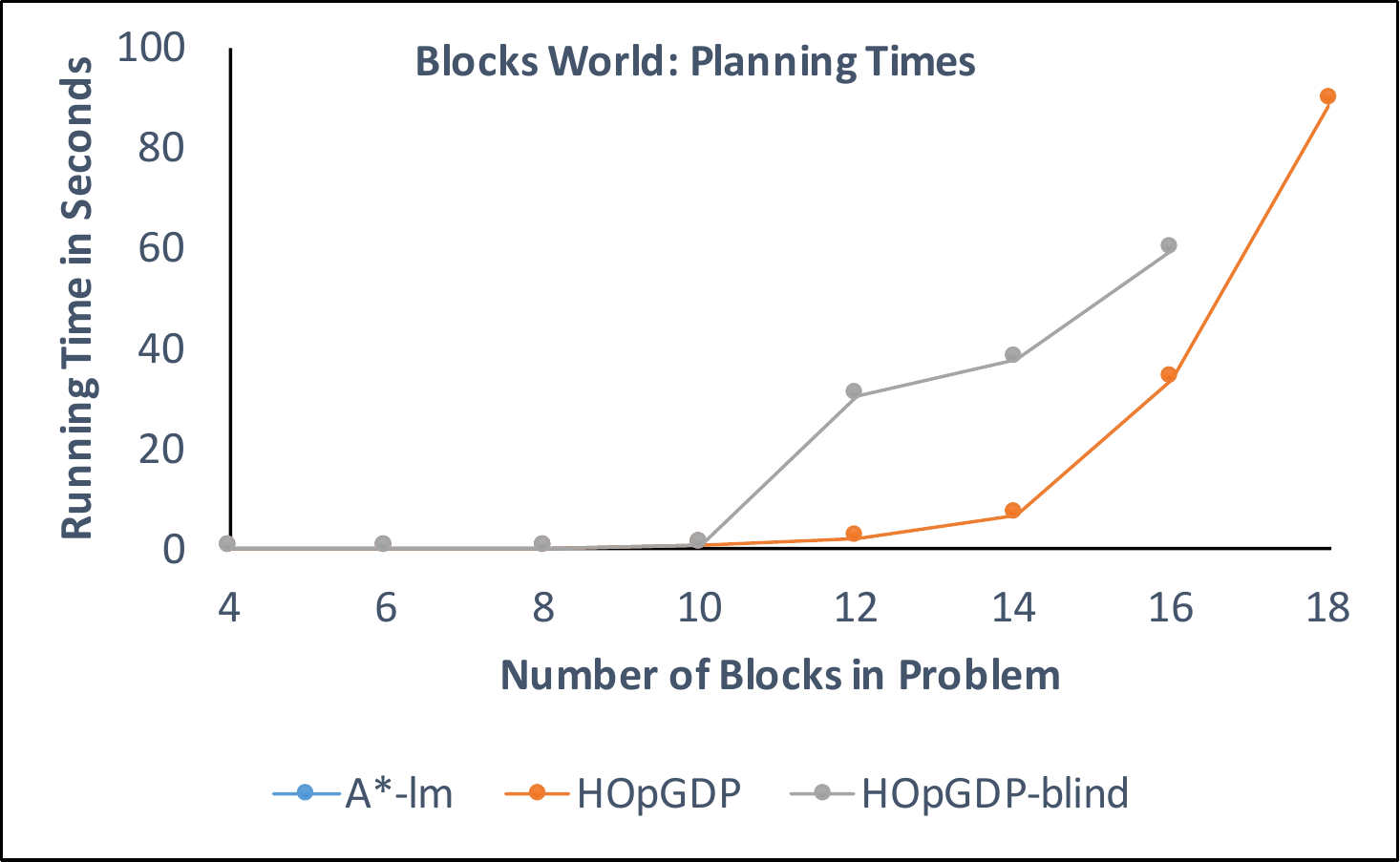}}
\subfloat[]{
            \label{fig:depots-running-time}
            \includegraphics[scale=0.40]{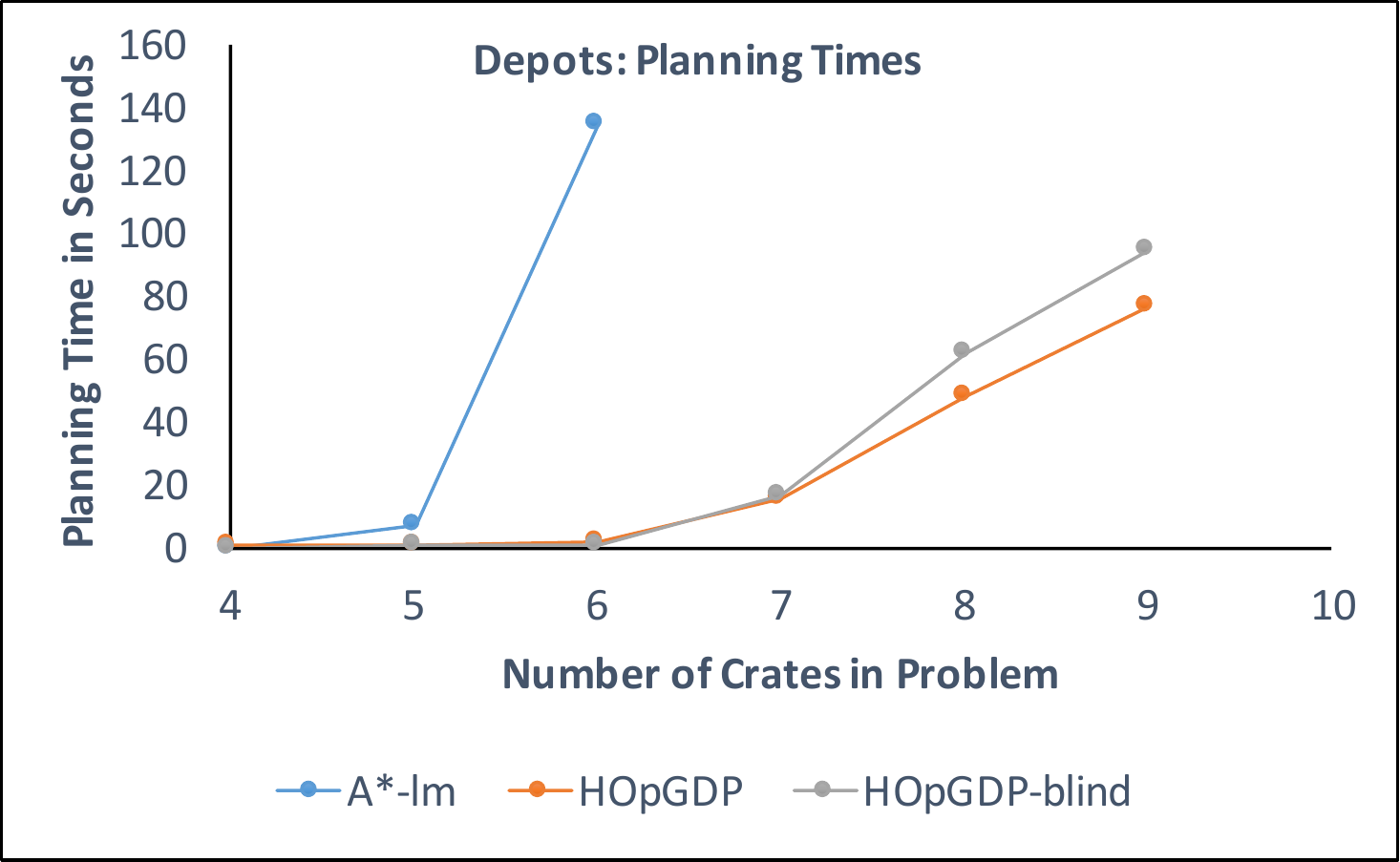}}
\caption{Graph of number of nodes expanded and running times of the planners across
the Logistics, Blocks-World, and Depots domains. Each data point is the
average over 25 randomly generated problems. Data points where
all the problems are not solved were discarded.}
    \label{fig:plots}
\end{figure*}
\section{Experimental Evaluation}
We implemented \hopgdp\ within the Fast-Downward codebase, and extended LAMA's landmark generation code to
develop \hgnlmheur, our HGN planning heuristic.

We tested two hypotheses in our study:
\begin{description}
    \item \textbf{H1}: \textit{\hopgdp's ability to exploit hierarchical planning knowledge enables it to
                outperform state-of-the-art optimal classical planners.}
                To test this, we compared the performances of \hopgdp\ with \astarlm~\cite{karpas09costoptimal}, the
                optimal classical planner whose heuristic we extended to develop \hgnlmheur.
                
                It might seem that \textbf{H1} is obviously true due to the dominance of hierarchical planners
                (e.g., SHOP2 and GDP) over classical planners,
                but these are merely satisficing planners.
                It is not clear whether this advantage would carry over to optimal planning
                because \hopgdp\ needs to do an optimal search in the possibly larger space of $(state,goal\text{-}network)$ pairs,
                in contrast to classical planners, which search in the space of states.

            \item \textbf{H2}: \textit{The heuristic used by \hopgdp, \hgnlmheur, provides useful
                search guidance.} To test this, we compared the performances of \hopgdp\ with \gdpblind, which is identical
                to \hopgdp\ except that it uses the trivial heuristic estimate of $h=0$.
\end{description}

\subsection{Experimental Results}
We evaluated \hopgdp, \gdpblind, and \astarlm\ on three well-known planning benchmarks,
Logistics, Blocks World and Depots.
We chose these 3 domains because from a control-knowledge standpoint, these three domains
capture a wide spectrum: Logistics contains only enough control-knowledge to define allowed
solutions, Blocks-World is at the other extreme, defining sophisticated knowledge that significantly
prunes the search space, and Depots incorporates elements of both.

For each domain, we randomly generated 25 problem instances per problem size.
We ran all problems on a Xeon E5-2639 with a per problem limit of 4 GB of RAM and 25 minutes of planning time.
Data points were discarded if the planner did not solve all of the corresponding problem instances within the
time limit.

\mypar{Logistics}
We modified the standard PDDL Logistics model
to limit the capacity of all
vehicles to one to ensure the HGN and non-HGN planners compute the same solutions.
We generated 25 random logistics problems for each problem size ranging from 4,6,\ldots,14 packages. For \hopgdp\ and
\gdpblind, we provided the HGN methods used in \godel's experimental evaluation~\cite{shivashankar2013godel}.
There are three methods in this knowledge base that together capture all the possible (minimal) solutions to a Logistics
problem; these are 
(1) a method to move packages within the same city using trucks,
(2) a method to move packages between airports using planes, and
(3) a method that combines the previous two to move packages across different cities.

Figures~\ref{fig:logistics-nodes-expanded} and \ref{fig:logistics-running-time} show the performance of the three planners in terms of
number of nodes expanded by the planners and overall planning time. Both \hopgdp\ and \gdpblind\ could solve problems up to size 10 (i.e., within the time limit), while \astarlm\ could solve problems only up to size 6. 

In terms of nodes expanded, Figure~\ref{fig:logistics-nodes-expanded} shows that the heuristic in \hopgdp\ helped to modestly decrease the number of nodes expanded; \hopgdp\ on average expanded 22\% fewer nodes than \gdpblind. We did not include \astarlm\ because it expanded many orders of magnitude more nodes than either \hopgdp\ variant (e.g., for problems of size 6, \astarlm\ on average expanded $12\times10^6$ nodes). With regard to running time, Figure~\ref{fig:logistics-running-time} shows that the modest gain by \hgnlmheur\ was outweighed by the computational overhead of running the heuristic (on average about 35\% of the total running time).
\gdpblind, despite its blind search, was slightly faster than \hopgdp. 

\mypar{Blocks World}
We generated 25 random blocks-world problems for problem sizes ranging from 4,6,\ldots,20 blocks. As in our study with Logistics,
we use the same HGN methods used in \godel's evaluation~\cite{shivashankar2013godel}. In contrast to our Logistics study,
the methods encode sophisticated knowledge that allows the planners to prune search paths that don't lead
to good solutions (e.g., it contains a recursively defined axiom that checks if a block is in its final position
and only then builds towers on top of it).

Figures~\ref{fig:bw-nodes-expanded} and \ref{fig:bw-running-time} show the performance of the three planners on these blocks-world problems. \astarlm\ could solve problems up to size 10, \gdpblind\ to size 16, and \hopgdp\ could solve problems up to size 18.

Figure~\ref{fig:bw-nodes-expanded} displays the number of nodes expanded by the three planners. In this domain, the
guidance provided by \hgnlmheur\ helped substantially;
\hopgdp\ on average expanded 76\% fewer nodes than \gdpblind. This savings far outweighed the heuristic computation overhead (on average about 48\% of the total running time), resulting in smaller overall planning times for \hopgdp\ as can be seen in Figure~\ref{fig:bw-running-time}.

\mypar{Depots}
We generated 25 random depots problems for problem sizes ranging from 4,5,\ldots,10 crates. Since the Depots domain
combines aspects of Logistics (moving cargo around) and Blocks-World (stacking them in a particular manner),
the HGN methods for Depots is a combination of the HGNs used in Logistics and Blocks-World.

Figures~\ref{fig:depots-nodes-expanded} and \ref{fig:depots-running-time} show the performance
of the three planners on the generated problems.
\astarlm\ could solve problems up to only size 6, while both \hopgdp\ and \gdpblind\ could solve problems
up to size 9.
Figure~\ref{fig:depots-nodes-expanded} shows the average number of nodes expanded by the three planners. The
\hgnlmheur\ heuristic in \hopgdp\ provides good search guidance, reducing the number of nodes expanded by
about 46\% when compared to \gdpblind. As in Logistics, we didn't show the nodes expanded by \astarlm\ since
it was many orders of magnitude more than either \hopgdp\ variant; for size 6 problems, on average, it expanded
$3.5\times10^6$ nodes.

In terms of planning time (Figure~\ref{fig:depots-running-time}), the provided
domain knowledge clearly helps both \hopgdp\ variants in scaling much better than \astarlm. Furthermore,
the additional search guidance provided by \hgnlmheur\ results in overall lower runtimes for \hopgdp\
in comparison to \gdpblind,
even with the computation overhead of the heuristic (which is about 56\% of the total time).

\subsection{Interpretation of Results}
\label{subsec:expt-takeaways}

There are two main takeaways from this empirical study:
\begin{description} \item[Support for H1.]
        \textit{Hierarchical planning knowledge helps in scaling up solving of optimal
        planning problems.} In all three benchmark domains, both of the \hopgdp\ variants solved more
        problems while requiring less time and expanding fewer nodes than \astarlm, showing that
        the additional overhead of searching through the space of $(state,goal\text{-}network)$
        pairs was
        outweighed by the benefit that hierarchical planning knowledge can provide in terms of
        more focused search.

    \item [Support for H2.]
        \textit{The HGN heuristic \hgnlmheur\ provides useful guidance when searching for hierarchically
        optimal plans.} We can conclude this from the decrease in the number of nodes expanded in
        \hopgdp\ as compared to \gdpblind\ in all three benchmark domains.
        The Logistics results only weakly support this due to only a modest decrease in the number
        of nodes expanded (22\%), while the results from Blocks-World and Depots
        are more conclusive, registering large savings in number of nodes expanded
        (76\% and 46\% respectively).

        We posit that the reduction in number of nodes expanded by 
         \hgnlmheur\ is a function of the input HGN knowledge.
        For instance, the Logistics methods
        do not encode any expert knowledge and instead only model the minimum knowledge required to capture
        the three ways to move a package: by truck, by plane, and by a combination of the two. Therefore,
        the goal networks always contain landmarks or more focused versions of landmarks
        (e.g.
        $\textsf{airplane-at}(ap_1,loc_1)$ instead of
        $\textsf{airplane-at}(ap_1,loc_1) \lor \textsf{airplane-at}(ap_2,loc_1)$)
        that can be detected by landmark generation algorithms.
        This means that
        the landmarks generated do not change much after a method
        application, implying that the heuristic estimates are unlikely to change much either.
        In contrast, methods in both Blocks-World and Depots contain specialized
        knowledge that, when applied, yield goal networks containing subgoals and orderings
        that cannot be detected by landmark generation algorithms. That is,
        when landmark generation is run on these goal networks, because the subgoals in the goal network serve
        as seeds for landmark generation, a richer set of landmarks will be generated, resulting in more informed heuristic estimates.
\end{description}

Another important takeaway from the experiments is the following:
\textit{the current implementation of \hgnlmheur\ imposes a substantial overhead on
        \hopgdp.} On average, it uses 35\%, 48\% and 56\% of the total planning time
        in Logistics, Blocks-World, and Depots respectively. This is partly due to the current
        implementation not being optimized. For instance, unlike landmark-based classical planners where
        the landmark graph needs to be computed only once for the final goal,
        \hopgdp\ needs to compute landmark
        graphs for every goal network it generates during search.
        Reusing the computed landmark graphs more effectively can potentially help in substantially
        reducing planning times.

\section{Related Work}


HTN planners solve planning problems by (1) forward state-space search, such as in the SHOP~\cite{nau99shop} and SHOP2~\cite{nau2003shop2} HTN planners, or (2) partial-order causal-link planning (POCL) techniques, such as in UMCP~\cite{erol94umcp} and in the hybrid planning literature~\cite{elkawkagy12improving,bercher2014hybrid}.

HGN planning can be translated to HTN planning in a plan-preserving manner \cite{alford2016hierarchical}, meaning we can, in theory, use any optimal HTN planner for optimal HGN planning.
However, there is little research on search heuristics for forward-search HTN planning \cite{alford2014feasibility,alford2016bound}. Therefore, planners often provide other domain-specific mechanisms for users to encode search strategies. For example, SHOP2 allows domain-specific knowledge, known as \textit{HTN methods}, to be specified in a 'good' order according to the user, and attempts to apply them in the same order. SHOP2 also provides support for external function calls~\cite{nau2003shop2} that can call arbitrary code to perform intensive computations, thus minimizing the choices that need to be made during search. For example, in the 2002 Planning Competition for hand-tailored planners, the authors of SHOP2 used a graph-algorithm library that SHOP2 could call externally to generate shortest paths~\cite{nau2003shop2}.


Waisbrot et al \cite{waisbrot2008combining} developed $H2O$, a HTN planner that augments SHOP2 with classical planning heuristics to make local decisions on which method to apply next by estimating how close the method's goal is to the current state. However, $H2O$ retains the depth-first search structure of SHOP2, making it difficult to generate high-quality plans.


Marthi et al~\cite{marthi2007angelic,marthi2008angelic} propose an HTN-like formalism called \textit{angelic hierarchical planning} that allows users to annotate abstract tasks with additional domain-specific information (i.e., lower and upper bounds on the costs of the possible plans they can be used to generate). They then use this information to compute hierarchically-optimal plans. In contrast, we require the costs of only the primitive actions and use domain-independent search heuristics to compute hierarchically-optimal plans.


There has been recent work on developing search heuristics for POCL HTN planners \cite{elkawkagy12improving,bercher2014hybrid}. However, these heuristics typically provide estimates on how many more plan refinement steps need to be taken from a search node to obtain a solution. This differs from plan quality estimates, which is our focus in this paper.


\textit{Hierarchical Goal Network} (HGN) Planning combines the hierarchical structure of HTN planning with the goal-based nature of classical planning. It therefore allows for easier infusion of techniques from classical planning into hierarchical planning, such as adapting the FF heuristic for \textit{method ordering} in the \gdp\ planner~\cite{shivashankar2012hierarchical}, and using landmark-based techniques to plan with partial amounts of domain knowledge in \godel~\cite{shivashankar2013godel}. However, both planners use depth-first search and inadmissible heuristics, so they cannot provide any guarantees of plan quality.


Another less-related domain-configurable planning formalism is \textit{Planning with Control Rules}~\cite{bacchus2000tlplan}, where domain-specific knowledge is encoded as \textit{linear-temporal logic} (LTL) formulas. TLPlan, one of the earliest planners developed under this formalism, used control rules written in LTL to prune trajectories deemed suboptimal by the user. There have also been attempts to develop heuristic search planners that can plan with LTL$_f$, a simplified version of LTL that works with finite traces.
This has been used to incorporate search heuristics to
solve for temporally extended goals written in LTL$_f$~\cite{baier2006planning}, planning for preferences \cite{sohrabiBaierMcIlraith09.ijcai.htnPlanning}, 
as well as to express landmark-based heuristics that guide classical planners~\cite{simon2015finding}.

\section{Conclusion}


Despite the popularity of hierarchical planning techniques in theory and practice, little effort has been devoted to developing domain-independent search heuristics that can provide useful search guidance towards high-quality solutions. As a result, end-users need to encode domain-specific heuristics into the domain models, which can make the domain-modeling process tedious and error-prone.


To address this issue, we leverage recent work on HGN planning, which allows tighter integration of hierarchical and classical planning, to develop (1) \hgnlmheur, an admissible HGN planning heuristic, and (2) \hopgdp, an A$^*$ search algorithm guided by \hgnlmheur\ to compute \textit{hierarchically-optimal} plans. Our experimental study showed that \hopgdp\ outperforms optimal heuristic search classical planners (due to its ability to exploit domain-specific planning knowledge) and optimal blind search HGN planners (due to the search guidance provided by \hgnlmheur).


There are several directions for future work, such as:
\begin{itemize}
   
\item \textbf{Extension to Anytime Planning}: 
    An obvious and a practically useful extension of this work is to
    extend \hopgdp\ to work in an \textit{anytime} manner (i.e.,
    generate a solution quickly such that a solution is available at
    any time during execution and then iteratively/continuously
    improve the plan's quality over time) instead of trying to compute the
    optimal solution up-front.
    We can of course adapt techniques used in
    anytime classical planners like LAMA, which runs a series of
    weighted-A$^*$ searches.
    However, we also plan to explore the use of \textit{block-deordering}~\cite{Siddiqui2015ContinuingPQ},
    a technique for continual plan improvement that seems to lend itself well to
    plans that are hierarchically structured.

\item \textbf{Extension to Temporal Planning}:
    We also plan on investigating temporal extensions of HGN planning and \hopgdp\ to
    develop search heuristics and hierarchical planners that can
    leverage procedural knowlege to find high-quality plans and
    schedules.

\end{itemize}

\begin{acknowledgements}
This work is sponsored in part by OSD ASD (R\&E). The information in this paper does not
necessarily reflect the position or policy of the sponsors, and no official endorsement should be inferred.
Ron Alford performed part of this work under an ASEE postdoctoral fellowship at NRL.
We also would like to thank the anonymous reviewers at ECAI 2016 for their insightful comments.
We would also like to thank the reviewers at HSDIP 2016 for useful feedback on a preliminary version of this paper.
\end{acknowledgements}

\bibliography{refs}
\bibliographystyle{ecai}

\end{document}